# Rule-Based Semantic Sensing

Przemyslaw Woznowski and Alun Preece

Cardiff University, School of Computer Science, 5 The Parade, Cardiff, UK
{p.r.woznowski,a.d.preece}@cs.cf.ac.uk

**Abstract.** Rule-Based Systems have been in use for decades to solve a variety of problems but not in the sensor informatics domain. Rules aid the aggregation of low-level sensor readings to form a more complete picture of the real world and help to address 10 identified challenges for sensor network middleware. This paper presents the reader with an overview of a system architecture and a pilot application to demonstrate the usefulness of a system integrating rules with sensor middleware.

## 1 Introduction & Motivation

A single sensor provides only partial information on the actual physical condition measured, e.g. an acoustic sensor only records audio signals. For an application to reason over sensor data, raw sensor readings have to be captured and often aggregated to form a more complete picture of the real-world condition measured. Sensor Network Middleware (SNM) aids this process. As defined in [1], "The main purpose of middleware for sensor networks is to support the development, maintenance, deployment, and execution of sensing-based applications". However, existing SNMs don't give the user – who can be an expert in some area that is not computer science – an opportunity to easily specify data aggregation logic themselves.

It is hypothesised that rules help to address this problem and can greatly improve the SNM. Moreover, such an approach to sensor networks addresses many of the 10 challenges for SNM, listed in [2], in the following way:

**Data Fusion** - Rules fuse simple facts to infer higher-level facts about the real world.
**Application Knowledge** - Expert's knowledge encoded into an automated system.
**Adaptability** - Applicable to any domain, non-programmers can write rules.
**Abstraction Support** - Each fact is an interpretation of data. How the data is interpreted is determined by an expert via rules.
**QoS Support** - Multiple combinations of rules and facts can often answer the same query. Solution can be explained by retracing the reasoning.

The remaining challenges: Network Heterogeneity, Dynamic Topology, Resource Constraints, Security and Scalability – need to be met by SNM. Additional benefits come from well-known advantages of using RBS systems: reproducibility, permanence, consistency, timeliness, efficiency, breadth, completeness, documentation, etc. – as identified in [3]. Finally, representing sensor data in the form of facts adds semantics. We propose the Rule-Based Semantic Sensor System (RBS3) which employs a Rule-Based System (RBS) on top of existing off-the-shelf SNM. The pilot application described in Section 3 was implemented to test the hypothesis that rules help to address the 10 Challenges and ease the development, maintenance, execution and extensibility of sensing-based applications.

## 2    Proposed System Architecture

The proposed system architecture in Figure 1 consists of four layers. The SNM layer serves as a bridge between physical sensors and the layer above it. It abstracts away the network heterogeneity by modelling hardware entities, and the output they produce, in software. The Interface layer is responsible for injecting sensor data, coming from the layer below it, into the Reasoning Engine layer in the form of facts. Its main function is to translate the SNMs output into facts, defined in terms of a semantic data model (for which we currently use RDF Schema for simplicity, although details of this are not included in this paper due to lack of space). The Reasoning Engine layer is the heart of the system. It continuously reasons over incoming facts and those already in the Working Memory (WM) to produce new, more complex facts. The more complex the facts, the higher the semantic enrichment and therefore more detailed picture of the real world. The Application layer bridges the user's interface with the Reasoning Engine. It exposes facts and queries, which persist in the KB, to the application. Moreover, it takes user's queries, pushes them to the layer below and returns the results in the format easily consumable by the application.

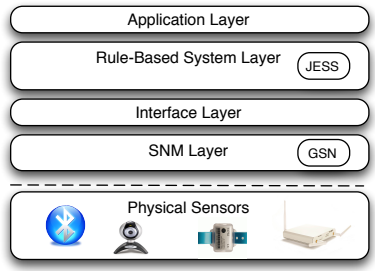

**Fig. 1.** System Architecture

The system architecture in Figure 1 is implemented in our RBS3 system as follows. The SNM layer currently consists of Global Sensor Networks (GSN) middleware, which serves XML data in response to queries. GSN (GNU GPL license) is a SNM, which deals with sensor network heterogeneity via use of Virtual Sensor (VS) abstraction. Any type of sensors, whether hardware or software, is represented by a single VS XML file, which specifies its inputs and output structure [4]. However, other SNMs such as: Pachube, ITA Sensor Fabric or SWE compliant middleware could replace GSN. The Interface Layer parses the XML data to JavaBeans, which are then injected into the Reasoning Engine (Jess) in the form of facts. Alternatively SweetRules, which is much more compact and offers extra features, could replace Jess, as both rule engines accept rules in CLIPS format. Queries, their arguments, and return parameters are available through the Application Layer, which serves data in JSON(JavaScript Object Notation) format, because it is a lightweight data-interchange format, which is easy for humans to read/write and easy for machines to parse.

## 3  Pilot Application

The aim of this application is to provide information on people's indoor locations, their history of visited locations, and information on walking speed between the locations - later referred in this paper as "corridor tests". The basic assumption for the system to work is that the tracked person wears either the RFID tag or any Bluetooth(BT) enabled device pre-registered with the mobileDevStore VS. Also corridor entities need to be defined in the corridorStore VS in order for the system to log corridor tests. This is part of a larger project looking at people's recovery from physical injury.

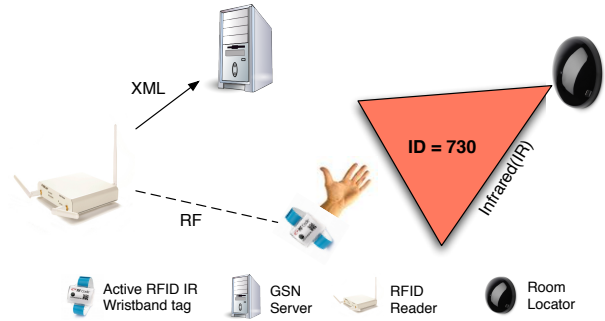

**Fig. 2.** Localisation with the Room Locator

The room-level localisation of active RFID tags is possible via use of the Room Locators, which broadcast a pre-set location code via IR (Figure 2). The active RFID wristband tags are IR enabled, therefore report the IR location code to the RFID reader. For this a direct line-of-sight between tag and the Room Locator is required. In the experiment, the network consisted of 1 laptop, 1 RFID reader, 2 Room Locators and 2 standard desktop PCs with Bluetooth, placed in two rooms, both running an instance of the GSN server. To clarify, the software/hardware components used in the experiment have the following functions:

**RFID Active Tag:** Every 2 seconds broadcasts it's unique ID, IR location code, motion status, etc. to the RFID Reader.

**Bluetooth:** Alternative source of information on user's location.
**GSN:** Connects to sensors, logs their readings and exposes them via a web server. Also serves as a source of information for static data.
**RFID Reader:** Receives active RFID tags' signals.
**Room Locator:** Transmits an IR pulse pattern containing a unique 3-digit location code to enable room-level accuracy localisation.
**mobileDevStore VS:** Lookup service. Stores name to RFID/BT address mappings.
**corridorStore VS:** Lookup service. Stores corridor entities (endA, endB, length).
**btReader VS:** Logs device's discovery time, BT address and reader's location.
**rfidReader VS:** Logs tag's discovery time, ID and reader's location.

### 3.1 Facts

Shadow fact, as described in [5], is "an unordered fact whose slots correspond to the properties of a JavaBean". Three shadow fact templates are defined in the Knowledge Base (KB): `MobileTrace`, `Person` and `Corridor`. They allow for quick insertion of JavaBean objects into the Working Memory (WM) and they directly represent GSN Virtual Sensor's outputs.

```
(deftemplate MobileTrace (declare (from-class javaBeans.MobileTrace)))
;Java class members/slots: location, address, time.
(deftemplate Person (declare (from-class javaBeans.Person)))
;Java class members/slots: name, deviceAddress.
(deftemplate Corridor (declare (from-class javaBeans.Corridor)))
;Java class members/slots: enda, endb, length.
```

Apart from shadow facts described above, the following set of unordered facts exists in the KB. All these facts originate from rules defined in the KB. To summarise, in this implementation, shadow facts (capitalised) represent sensor readings and unordered facts are used internally in Jess to represent fused sensor data. These fact templates are the semantic interface and we do have the RDF Schema for them, however, this is not included due to lack of space.

```
(deftemplate is-seen-at (slot name)(slot location)(slot time))
(deftemplate is-currently-at (slot name)(slot location)(slot tStart)
 (slot tFinish))
(deftemplate was-at(slot name)(slot location)(slot tStart)(slot tFinish))
(deftemplate was-tracked (slot name) (slot endA) (slot endB)
 (slot tStart)(slot tFinish)(slot distance)(slot tTaken)(slot velocity))
```

### 3.2 Rules

The set of rules defined in the KB, allows the system to infer four types of observations from sensor and static data: `is-seen-at`, `is-currently-at`, `was-at` and `was-tracked`. First rule, seen_at, simply aggregates `Person` and `MobileTrace` facts to assert `is-seen-at` to the WM. It also retracts all the `MobileTraces` that are successfully fused with `Person` facts.

```
(defrule seen_at
 (Person (deviceAddress ?address)(name ?name))
 ?mob <- (MobileTrace (location ?loc)(time ?time)(address ?address))
 =>  (retract ?mob)
     (assert (is-seen-at(name ?name)(location ?loc)(time ?time))))
```

The next rule, was_at, asserts two facts to the WM: was-at and is-currently-at. The latter contains information about a person's current location; therefore whenever the same person is seen at different location, the is-currently-at fact becomes was-at and a new is-currently-at fact is added. This time both facts, which are used to infer new information (is-seen-at and is-currently-at) are retracted from the WM, as at any point in time there should only exist one of each of these facts, simply because some person can only be seen at one location at any time. However, was_at will never fire unless the initial is-currently-at fact is inserted as is-currently-at facts are only produced by this rule. Therefore, a dummy fact is defined for each person tracked by the system, e.g. for Pete we have (assert(is-currently-at(name ''Pete'')(location ''dummyLoc'')(tStart 0)(tFinish 0))).

```
(defrule was_at
 ?c <- (is-currently-at(name ?n)(location ?l1)(tStart ?tS)(tFinish ?tF))
 ?seen <- (is-seen-at (name ?n)(location ?l2)(time ?t))
 =>(retract ?c ?seen)
   (assert(was-at(name ?n)(location ?l1)(tStart ?tS)(tFinish ?tF)))
   (assert(is-currently-at(name ?n)(location ?l2)(tStart ?t)(tFinish ?t))))
```

As opposed to was_at, the update_current_loc rule deals with the situation when the location reported by is-seen-at is the same: it simply updates the tFinish of the is-currently-at fact.

```
(defrule update_current_loc
 ?c <- (is-currently-at (name ?n)(location ?loc)(tStart ?tS)(tFinish ?tF))
 ?seen <- (is-seen-at (name ?n)(location ?loc)(time ?time))
 (test(< ?tF ?time))
 => (retract ?seen)(modify ?c (tFinish ?time)))
```

The three rules discussed above can already provide information on a subject's current location and history of visited locations. If a human expert was to analyse this data, they could easily answer questions on where the person currently is/was at any point in time. Additionally, it wouldn't be a problem to tell how much time it took somebody to transfer from one location to another, as this can be worked out from was-at facts. Find_corridor_events does exactly this, but in a slightly different way. Instead of analysing consecutive was-at facts it works with is-currently-at and was-at facts, whose locations are defined as ends of some Corridor in the KB. However, was-tracked fact is asserted if the subject travels from A to B and not B to A. This logic is there in purpose, as one may be interested in journeys in only one direction.

```
(defrule find_corridor_events
 (was-at (name ?name)(location ?loc1)(tStart ?t1S)(tFinish ?t1F))
 (is-currently-at (name ?name)(location ?loc2)(tStart ?t2S)(tFinish ?2tF))
 (Corridor (enda ?loc1)(endb ?loc2)(length ?length))
 => (bind ?tTaken (- ?t2S ?t1F))
    (assert (was-tracked (name ?name)(endA ?loc1)(endB ?loc2)(tStart ?t1F)
    (tFinish ?t2S)(distance ?length)(tTaken ?tTaken)
    (velocity (/ ?length (/ ?tTaken 1000)))))
```

It seemed to be enough to define only four rules in the KB. However, test results have revealed the missing logic. Assuming a scenario where somebody visits locations in the following order: 730, 000, 740, 000, 730, 000, 740 and the corridor is defined as (Corridor (enda 730)(endb 740)(length 20)) any person would know that there are two journeys of interest: 2 x "730 through 000 to 740". However, the system inferred one additional fact: "730 through 000, 740, 000, 730, 000 to 740". Since it does not make sense to consider cyclic journeys, we also have rules to retract these from the WM. Obviously the `find_corridor_events` rule could be replaced with a query, which looks for `was-at` and `Corridor` facts, however the general idea is to infer new, more complex facts from existing lower-level facts, rather that to come up with a clever query which can provide information on one's journeys. By inserting new and often more complex facts, the KB is populated with more data what allows for defining new rules that can simply look at existing facts and infer even more complex ones. `Find_corridor_events` is an example of a rule that does not modify facts that are already in the WM but instead populates new facts, which can then be used by other rules.

### 3.3 Results

Three queries, that take name as the parameter, are defined in the KB: `find_journeys`, `where_is` and `location_history`. They simply look for `was-tracked`, `is-currently-at` and `was-at` facts respectively for some person. Querying the WM becomes very simple, as neither new data needs to be inferred, nor any calculations done - simply query parameter needs defining. Hence query of the form `"find_journeys Pete"` lists all the `was-tracked` facts (corridor test results) for Pete.

To validate this application some tests were carried out. The table below contains results of the corridor tests recorded by the system, contrasted with times recorded by the subject of these tests via use of an ordinary watch synchronised with system's time. For simplicity, times represented in the table are of form HH:MM:SS and do not include milliseconds. From Table 1 it is easy to see that the system never underestimates the

**Table 1.** Experiment Results

| Recorded by the system | | | Recorded by hand | | |
|---|---|---|---|---|---|
| tStart | tFinish | tTaken | tStart | tFinish | tTaken |
| 13:30:44 | 13:31:26 | 42 | 13:30:43 | 13:31:21 | 38 |
| 13:36:21 | 13:37:00 | 39 | 13:36:18 | 13:36:56 | 38 |
| 13:59:25 | 14:00:08 | 43 | 13:59:22 | 14:00:03 | 41 |
| 14:13:41 | 14:14:16 | 35 | 14:13:38 | 14:14:08 | 30 |

tTaken but is rather an overestimate of it. This behaviour was predictable due to the following two factors. Firstly, RFID tags broadcast their signal every 2 seconds (when in motion) and therefore introduce a maximum delay of 2 seconds on both ends of the corridor. Therefore, if somebody arrives at some location, this information may not be injected into the system until the next round of broadcasting, which in worst case is 2 seconds later. Secondly, in order for the tag to report it's location it has to receive the IR signature of some location. If there is no direct line-of-sight between the tag and Room Locator, the tag reports location code 000 instead of the broadcasted location code. To account for both these factors the system could subtract the average delay time from the results returned.

## 4 Related Work

Many of the popular Sensor Network Middlewares, such as GSN, ITA Sensor Fabric, Pachube or SWE-compatible products are rather low-level [4,6,7,8]. They simply pro-vide sensor data (using different models and abstractions) and do not make it very easy for the programmer to program with them. We consider these as candidates for the SNM layer rather than complete solutions that meet all the 10 challenges to a satisfactory level. Application knowledge, adaptability and abstraction support are not very well addressed by these SNMs.

To the best of our knowledge there are no systems that implement a Rule-Based System on the top of SNM. The most similar is Semantic Streams (SS) – "...a framework ...that allows users to pose declarative queries over semantic interpretations of sensor data" [9]. SS and RBS3 are both very high-level in terms of ability to query for real world facts. RBS3 adapts ideas from SS in a sense that rules have analogous function to the semantic services – both take some inputs and produce outputs as a result of data aggregation. Moreover, streams of data (in case of RBS3 - facts) are reused in both systems. However, SS uses a modified version of Prolog and connects to sensors using MSR Sense toolkit, hence lacks openness at the lower layer, and can only use sensors compatible with this toolkit – according to Microsoft [10] "MSR Sense has only been tested on TinyOS-based sensor motes, although in theory, it should work with any 802.15.4 compatible wireless sensors". Therefore SS is hard to extend with new sensors or other sensor middleware. In addition rules are coded implicitly using SS markup language another specification to learn in order to use the system. RBS3, on the other hand, defines rules explicitly in a well-known "standard" form of rule (CLIPS) and allows adding new rules at the runtime. Semantic Streams use a variant of backward chaining to find semantic services that can satisfy the query. In contrast, RBS3 implements forward-chaining mechanism and only allows the user to query the system using queries defined in the KB.

## 5 Conclusion & Future Work

In this paper, we have proposed a system architecture which combines rules and sensor middleware to better address 10 identified challenges for sensing systems. The proposed system architecture provides several benefits amongst which are: flexibility and extensibility. This approach also aids application development, maintenance, deployment, and execution. Other benefits come from using a Rule-Based System and they help to address half of the 10 challenges for SNM: Data Fusion, Application Knowledge, Adaptability, QoS and Abstraction Support.

The current implementation of the system only has GSN in the SNM layer. In the next version of RBS3, wrappers to interface with other popular sensor middleware, such as Pachube, ITA Sensor Fabric or SWE, will be present. The proposed system architecture makes the system extensible – if the user is constrained to use a specific type of SNM they can implement their own wrapper for it; and flexible – if the user does not want to be limited to use one SNM but wants to use sensor data from various sources. Another improvement to the system would be to modify the Interface Layer, so that RDF data serialised in JSON is parsed and injected into the Reasoning Engine, instead of XML parsed to JavaBeans – "since XML just describes grammars there is no way of recognising a semantic unit from a particular domain of interest" [11]. The Reasoning Engine would then be processing semantically rich data.

Because the system works in a forward-chaining way, only when a rule that produces certain type of facts is specified, these facts become available for queries. The next version of the system may use both: backward- and forward-chaining mechanisms to allow the user to query for data for which production rules are specified just before the

query – therefore, historical data stored in DBs can participate to the query. Another, desirable enhancement to the system is at the Application layer – the system is easier to interface with if the user has a choice whether to receive data in JSON or RDF format.

Scalability is something that the entire system, once fully implemented, has to be extensively tested for in order to provide good response times and good level of reliability – the more sensors used, the more data to parse and store. However, the system as it is, is proven to work correctly and starts to reveal it's potential. As facts are injected into the system, they are not only aggregated together but also they are re-used across multiple rules. The more complex the facts are the better they re-create the real world conditions measured by sensors and can answer more complex queries.

## Acknowledgements

We would like to thank Prof. Robert van Deursen, Arshi Iqbal (School of Healthcare Studies, Cardiff University) and Dr Allison Cooper (Swansea University) for providing RFID equipment and continuous help with system development and testing. Many thanks to Chris Gwilliams for proofreading this paper.